\documentclass[11pt, a4paper, logo, copyright, nonumbering]{deepseek}
\usepackage[numbers, sort&compress]{natbib}

\usepackage{dblfloatfix}
\usepackage{ulem}
\usepackage{caption}
\usepackage{dramatist}
\usepackage{xspace}
\usepackage{pifont}
\usepackage{multirow}
\usepackage{tcolorbox}
\usepackage{xltabular}
\usepackage{longtable}
\usepackage{pdfpages}
\definecolor{linkcolor}{rgb}{0.956,0.298,0.235}
\definecolor{citecolor}{HTML}{1976D2}
\usepackage{tabularx}
\usepackage{hyperref}
\usepackage{graphicx}
\usepackage{subfigure}
\usepackage{amsmath}
\usepackage{amssymb}
\usepackage{array}
\usepackage{booktabs}
\usepackage{multirow}
\usepackage{arydshln}
\usepackage[utf8]{inputenc}
\hypersetup{
    colorlinks=true,
    linkcolor=citecolor,
    filecolor=magenta,
    urlcolor=cyan,
    citecolor=citecolor,
}

\usepackage{amsfonts}
\usepackage{amsmath}
\usepackage{amssymb}
\usepackage{lineno}

\usepackage[bottom]{footmisc}

\usepackage{CJKutf8}
\usepackage{subfigure}
\usepackage{setspace}

\usepackage{xspace}

\makeatletter
\def\@BTrule[#1]{
  \ifx\longtable\undefined
    \let\@BTswitch\@BTnormal
  \else\ifx\hline\LT@hline
    \nobreak
    \let\@BTswitch\@BLTrule
  \else
     \let\@BTswitch\@BTnormal
  \fi\fi
  \global\@thisrulewidth=#1\relax
  \ifnum\@thisruleclass=\tw@\vskip\@aboverulesep\else
  \ifnum\@lastruleclass=\z@\vskip\@aboverulesep\else
  \ifnum\@lastruleclass=\@ne\vskip\doublerulesep\fi\fi\fi
  \@BTswitch}
\makeatother

\addto\extrasenglish{

}

\newlength\savewidth\newcommand\shline{\noalign{\global\savewidth\arrayrulewidth
  \global\arrayrulewidth 1pt}\hline\noalign{\global\arrayrulewidth\savewidth}}

\reportnumber{001}

\renewcommand{\today}{}

\title{\centering \includegraphics[scale=0.05, bb=0 166 200 0]{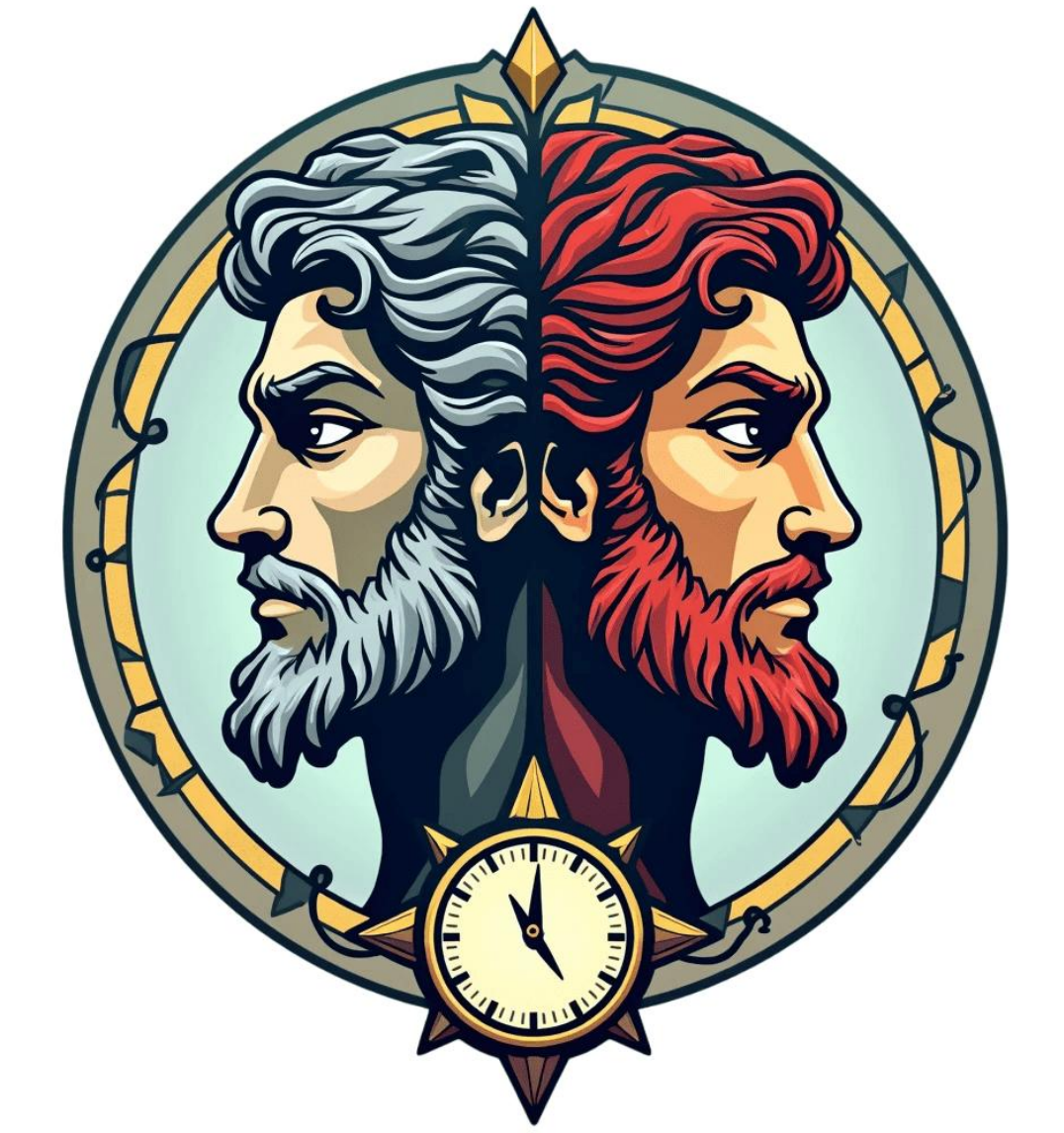} \quad \  Janus-Pro: Unified Multimodal Understanding and Generation with Data and Model Scaling}

\author[*]{
\small

Xiaokang Chen, Zhiyu Wu, Xingchao Liu, Zizheng Pan, Wen Liu, Zhenda Xie, Xingkai Yu, Chong Ruan

\small
DeepSeek-AI \\
\small
Project Page: \url{https://github.com/deepseek-ai/Janus}

}

\begin{abstract}
In this work, we introduce \textbf{Janus-Pro}, an advanced version of the previous work Janus. Specifically, Janus-Pro incorporates (1) an optimized training strategy, (2) expanded training data, and (3) scaling to larger model size. 
With these improvements, Janus-Pro achieves significant advancements in both multimodal understanding and text-to-image instruction-following capabilities, while also enhancing the stability of text-to-image generation. We hope this work will inspire further exploration in the field. Code and models are publicly available.

\end{abstract}

\begin{document}
\begin{CJK*}{UTF8}{gbsn}

\maketitle

\section{Introduction}

\begin{figure}[ht]
\begin{center}
    \subfigure[Average performance on four multimodal understanding benchmarks.]
    {
        \includegraphics[width=0.48\linewidth]{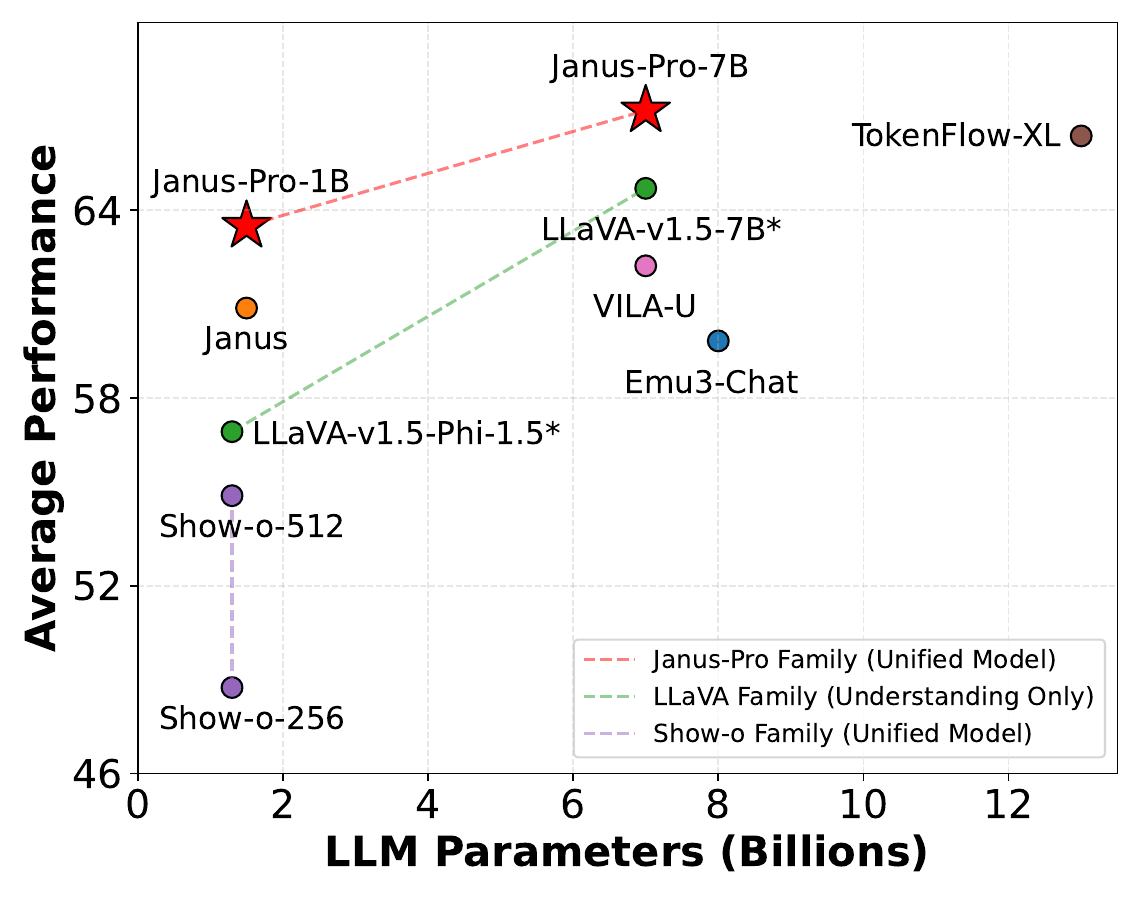}
        \label{fig:teaser_understanding}
    }
    \hfill
    \subfigure[Performance on instruction-following benchmarks for text-to-image generation.]
    {
        \includegraphics[width=0.48\linewidth]{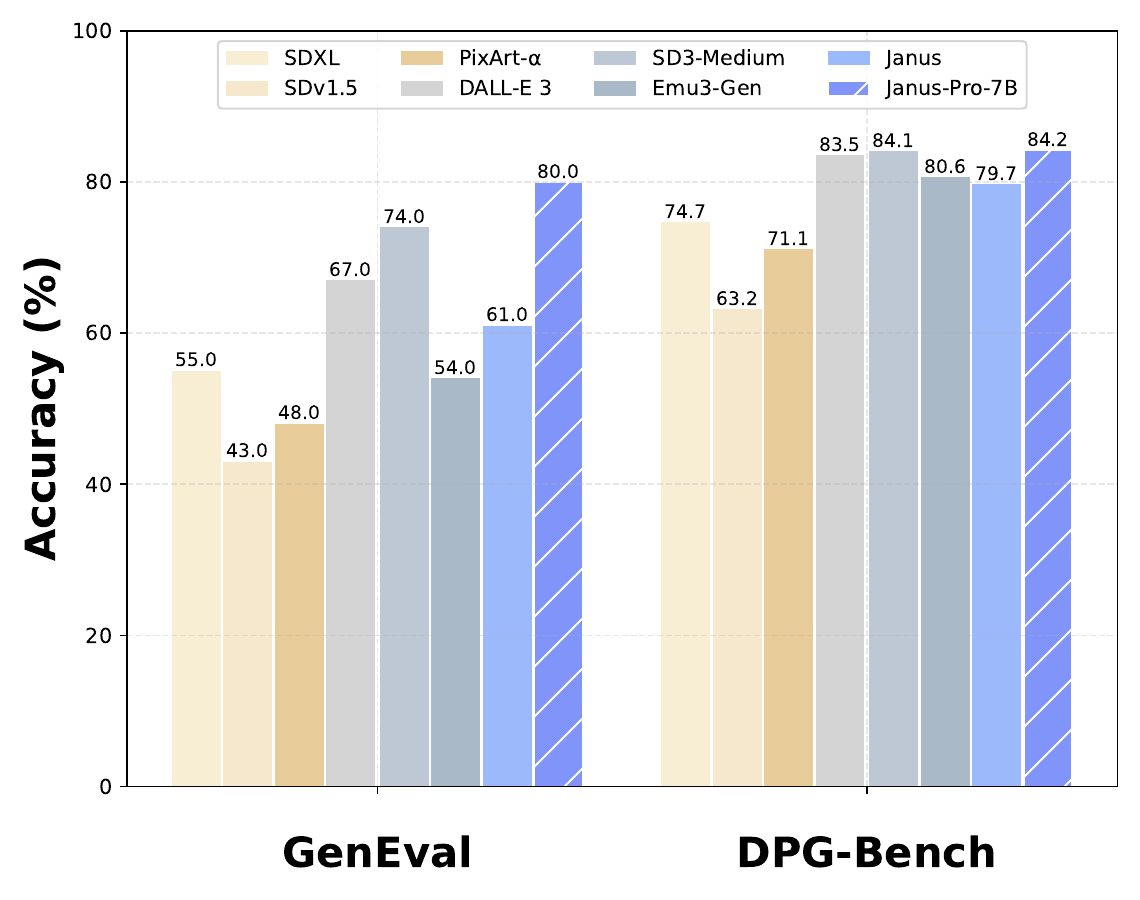}
        \label{fig:teaser_generation}
    }
\end{center}
\caption{\small
\textbf{Multimodal understanding and visual generation results from our Janus-Pro}. 
For multimodal understand, we average the accuracy of POPE, MME-Perception, GQA, and MMMU. The scores of MME-Perception are divided by 20 to scale to $[0, 100]$. For visual generation, we evaluate the performance on two instruction-following benchamrks, GenEval and DPG-Bench. Overall, Janus-Pro outperforms the previous state-of-the-art unified multimodal models as well as some task-specific models. Best viewed on screen.
}
\label{fig:teaser}
\end{figure}

\begin{figure}[ht]
    \centering
    \includegraphics[width=\textwidth]{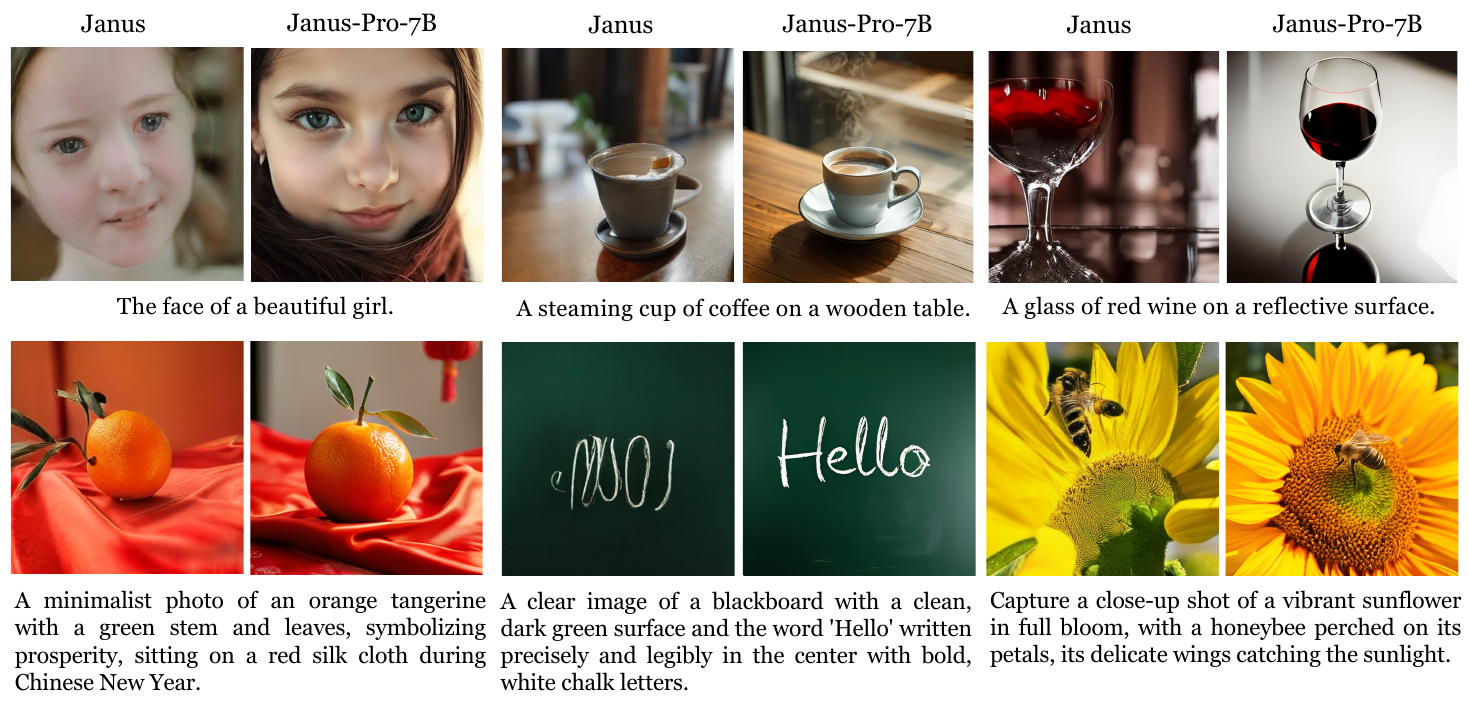}
    \caption{\textbf{Comparison of text-to-image generation between Janus-Pro and its predecessor, Janus.} Janus-Pro delivers more stable outputs for short prompts, with improved visual quality, richer details, and the ability to generate simple text. The image resolution is $384 \times 384$. Best viewed on screen.    } 
    \label{fig:compare}
\end{figure}

Recent advancements in unified multimodal understanding and generation models have demonstrated significant progress~\cite{team2024chameleon,wu2024vila,xie2024show,zhou2024transfusion,wu2024janus,ma2024janusflow,wang2024emu3,zhao2024monoformer}. These approaches have been proven to enhance the instruction-following capabilities in visual generation tasks while reducing model redundancy. 
Most of these methods utilize the same visual encoder to process inputs for both multimodal understanding and generation tasks. Since the representations required for these two tasks differ, this often results in suboptimal performance in multimodal understanding. To address this issue, Janus~\cite{wu2024janus} proposes decoupling visual encoding, which alleviates the conflict between multimodal understanding and generation tasks, achieving excellent performance in both tasks.

As a pioneering model, Janus is validated at the 1B parameter scale. However, due to the limited amount of training data and the relatively small model capacity, it exhibites certain shortcomings, such as suboptimal performance on short prompts image generation and unstable text-to-image generation quality. In this paper, we introduce Janus-Pro, an enhanced version of Janus that incorporates improvements across three dimensions: training strategies, data, and model size. The Janus-Pro series includes two model sizes: 1B and 7B, demonstrating scalability of the visual encoding decoding method.

We evaluate Janus-Pro on multiple benchmarks, and the results reveal its superior multimodal understanding capabilities and significantly improved text-to-image instruction-following performance. Specifically, Janus-Pro-7B achieved a score of 79.2 on the multimodal understanding benchmark MMBench~\cite{liu2023mmbench}, surpassing state-of-the-art unified multimodal models such as Janus~\cite{wu2024janus} (69.4), TokenFlow~\cite{qu2024tokenflow} (68.9) and MetaMorph~\cite{tong2024metamorph} (75.2). Additionally, in the text-to-image instruction-following leaderboard GenEval~\cite{ghosh2024geneval}, Janus-Pro-7B scores 0.80, outperforming Janus~\cite{wu2024janus} (0.61), DALL-E 3 (0.67), and Stable Diffusion 3 Medium~\cite{esser2024scalingrectifiedflowtransformers} (0.74).

\begin{figure}[ht]
    \centering
    \includegraphics[width=\textwidth]{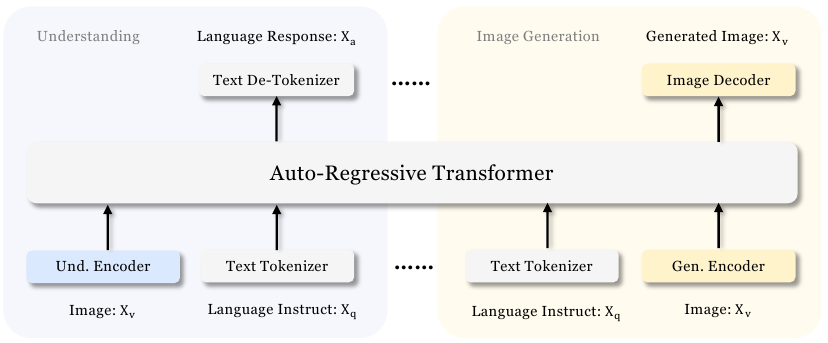}
    \caption{\textbf{Architecture of our Janus-Pro.} We decouple visual encoding for multimodal understanding and visual generation. ``Und. Encoder'' and ``Gen. Encoder'' are abbreviations for ``Understanding Encoder'' and ``Generation Encoder'', respectively. Best viewed on screen.    } 
    \label{fig:architecture}
\end{figure}

\section{Method}

\subsection{Architecture}
The architecture of Janus-Pro is shown in Figure~\ref{fig:architecture}, which is the same as Janus~\cite{wu2024janus}. The core design principle of the overall architecture is to decouple visual encoding for multimodal understanding and generation.
We apply independent encoding methods to convert the raw inputs into features, which are then processed by an unified autoregressive transformer. For multimodal understanding, we use the SigLIP~\cite{zhai2023sigmoid} encoder to extract high-dimensional semantic features from images. These features are flattened from a $2$-D grid into a $1$-D sequence, and an understanding adaptor is used to map these image features into the input space of the LLM.
For visual generation tasks, we use the VQ tokenizer from~\cite{sun2024autoregressive} to convert images into discrete IDs. After the ID sequence is flattened into $1$-D, we use a generation adaptor to map the codebook embeddings corresponding to each ID into the input space of the LLM. We then concatenate these feature sequences to form a multimodal feature sequence, which is subsequently fed into the LLM for processing. Apart from the built-in prediction head in the LLM, we also utilize a randomly initialized prediction head for image predictions in the visual generation task. The entire model adheres to an autoregressive framework.

\subsection{Optimized Training Strategy}

The previous version of Janus employs a three-stage training process. Stage I focuses on training the adaptors and the image head.
Stage II handles unified pretraining, during which all components except the understanding encoder and the generation encoder has their parameters updated.
Stage III is supervised fine-tuning, building upon Stage II by further unlocking the parameters of the understanding encoder during training. This training strategy has certain issues.
In Stage II, Janus  divides the training for text-to-image capabilities into two parts following PixArt~\cite{chen2023pixart}. The first part trains on ImageNet~\cite{deng2009imagenet} data, using image category names as prompts for text-to-image generation, with the goal of modeling pixel dependence. The second part trains on normal text-to-image data. During implementation, 66.67\% of the text-to-image training steps in Stage II are allocated to the first part. However, through further experimentation, we find that this strategy is suboptimal and lead to significant computational inefficiency.

To address this issue, we make two modifications.
\begin{itemize}
    \item \textbf{Longer Training in Stage I}: We increase the training steps in Stage I, allowing sufficient training on the ImageNet dataset. Our findings reveals that even with the LLM parameters fixed, the model could effectively model pixel dependence and generate reasonable images based on category names.
    \item \textbf{Focused Training in Stage II}: In Stage II, we drop ImageNet data and directly utilize normal text-to-image data to train the model to generate images based on dense descriptions. This redesigned approach enables Stage II to utilize the text-to-image data more efficiently, resulting in improved training efficiency and overall performance.
\end{itemize}

We also adjust the data ratio in Stage III supervised fine-tuning process across different types of datasets, changing the proportion of multimodal data, pure text data, and text-to-image data from 7:3:10 to 5:1:4. By slightly reducing the proportion of text-to-image data, we observe that this adjustment allows us to maintain strong visual generation capabilities while achieving improved multimodal understanding performance.

\subsection{Data Scaling}
We scale up the training data used for Janus in both multimodal understanding and visual generation aspects.

\begin{itemize}
\item \textbf{Multimodal Understanding}. For the Stage II pretraining data, we refer to DeepSeek-VL2~\cite{wu2024deepseek} and add approximately 90 million samples. These include image caption datasets (e.g., YFCC~\cite{mehdidc_yfcc_15m}), as well as data for table, chart, and document understanding (e.g., Docmatix~\cite{docmatix}). For the Stage III supervised fine-tuning data, we also incorporate additional datasets from DeepSeek-VL2, such as MEME understanding, Chinese conversational data, and datasets aimed at enhancing dialogue experiences. These additions significantly expanded the model's capabilities, enriching its ability to handle diverse tasks while improving the overall conversational experience.
\item \textbf{Visual Generation}. We observe that the real-world data used in the previous version of Janus lacks quality and contains significant noise, which often leads to instability in text-to-image generation, resulting in aesthetically poor outputs. In Janus-Pro, we incorporate approximately 72 million samples of synthetic aesthetic data, bringing the ratio of real to synthetic data to 1:1 during the unified pretraining stage. The prompts for these synthetic data samples are publicly available, such as those in~\cite{vivym2023midjourneyprompts}. Experiments demonstrat that the model converges faster when trained on synthetic data, and the resulting text-to-image outputs are not only more stable but also exhibit significantly improved aesthetic quality.
\end{itemize}

\subsection{Model Scaling}
The previous version of Janus validates the effectiveness of visual encoding decoupling using a 1.5B LLM. In Janus-Pro, we scaled the model up to 7B, with the hyperparameters of both the 1.5B and 7B LLMs detailed in Table~\ref{tab:arch}. We observe that when utilizing a larger-scale LLM, the convergence speed of losses for both multimodal understanding and visual generation improved significantly compared to the smaller model. This finding further validates the strong scalability of this approach.

\begin{table}[t!]
\caption{
\textbf{Architectural configuration for Janus-Pro}. We list the hyperparameters of the architecture.
}
\centering
\renewcommand{\arraystretch}{1.2}
\scalebox{0.9}{
\begin{tabular}{l|c|c}
    \shline
    & \textbf{Janus-Pro-1B} & \textbf{Janus-Pro-7B} \\
    \shline
    Vocabulary size & 100K & 100K  \\
    Embedding size & 2048 & 4096 \\
    Context Window & 4096 & 4096  \\
    \#Attention heads & 16 & 32  \\
    \#Layers & 24 & 30  \\
    \shline
    
\end{tabular}
}
\label{tab:arch}
\end{table}

\begin{table}[ht]
\caption{
\textbf{Detailed hyperparameters for training Janus-Pro}. Data ratio refers to the ratio of multimodal understanding data, pure text data, and visual generation data.
}
\centering
\setlength{\tabcolsep}{2pt}
\renewcommand{\arraystretch}{1.2}
\scalebox{0.9}{
\begin{tabular}{l|ccc|ccc}
\shline
&  \multicolumn{3}{c}{Janus-Pro-1B} \vline & \multicolumn{3}{c}{Janus-Pro-7B} \\
\shline
\textbf{Hyperparameters} & \textbf{Stage 1} & \textbf{Stage 2}         & \textbf{Stage 3} & \textbf{Stage 1} & \textbf{Stage 2}  & \textbf{Stage 3}  \\
\hline
Learning rate         & $1.0\times10^{-3}$   & $1.0\times10^{-4}$ & $4.0\times10^{-5}$& $1.0\times10^{-3}$   & $1.0\times10^{-4}$ & $4.0\times10^{-5}$  \\
LR scheduler       & Constant   & Constant & Constant & Constant  & Constant & Constant  \\
Weight decay      & 0.0   & 0.0 & 0.0 & 0.0 & 0.0 & 0.0  \\
Gradient clip       & 1.0 & 1.0 & 1.0 & 1.0 & 1.0 & 1.0  \\
Optimizer      & \multicolumn{3}{c}{AdamW ($\beta_1=0.9, \beta_2=0.95$)} \vline & \multicolumn{3}{c}{AdamW ($\beta_1=0.9, \beta_2=0.95$)}  \\
Warm-up steps        & 600 & 5000 & 0 & 600 & 5000 & 0  \\
Training steps        & 20K & 360K & 80K & 20K & 360K & 40K  \\
Batch size        & 256 & 512 & 128 & 256 & 512 & 128  \\
Data Ratio        & 1:0:3 & 2:3:5 & 5:1:4 & 1:0:3 & 2:3:5 & 5:1:4  \\
\shline
\end{tabular}}
\label{tab:hyper}
\end{table}

\section{Experiments}

\subsection{Implementation Details}

In our experiments, we utilize DeepSeek-LLM ($1.5$B and $7$B) \cite{bi2024deepseek} with a maximum supported sequence length of $4096$ as the base language model. For the vision encoder used in understanding tasks, we select SigLIP-Large-Patch$16$-$384$ \cite{zhai2023sigmoid}. The generation encoder has a codebook of size $16,384$ and downsamples images by a factor of $16$. Both the understanding adaptor and the generation adaptor are two-layer MLPs. The detailed hyperparameters for each stage are provided in Table~\ref{tab:hyper}. Please note that for Stage II, we employ an early stopping strategy, halting at 270K steps. All images are resized to $384 \times 384$ pixels. 
For multimodal understanding data, we resize the long side of the image and pad the short side with the background color (RGB: $127$, $127$, $127$) to reach $384$. 
For visual generation data, the short side is resized to $384$, and the long side is cropped to $384$. We use sequence packing during training to improve training efficiency. We mix all data types according to the specified ratios in a single training step. Our Janus-Pro is trained and evaluated using HAI-LLM~\cite{haillm}, which is a lightweight and efficient distributed training framework built on top of PyTorch. The whole training process took about $9$/$14$ days on a cluster of $16$/$32$ nodes for $1.5$B/$7$B model, each equipped with $8$ Nvidia A$100$ ($40$GB) GPUs.

\begin{table}[ht]
    \centering
    \setlength{\tabcolsep}{1.05pt}
    \renewcommand{\arraystretch}{1.2}
    \scriptsize
    \caption{\textbf{Comparison with state-of-the-arts on multimodal understanding benchmarks}. ``Und.'' and ``Gen.'' denote ``understanding'' and ``generation'', respectively. Models using external pretrained diffusion model are marked with $^\dagger$.}
    \label{sota_result_understanding}
    \begin{tabular}{llcccccccc}
        \toprule
        \textbf{Type} & \textbf{Model} & \textbf{\# LLM Params} & \textbf{POPE$ \uparrow$} & \textbf{MME-P$ \uparrow$} & \textbf{MMB$ \uparrow$} & \textbf{SEED$ \uparrow$}  & \textbf{GQA$ \uparrow$} & \textbf{MMMU$ \uparrow$} & \textbf{MM-Vet$ \uparrow$} \\
        \midrule
        \textit{Und. Only} & 
        LLaVA-v$1.5$-Phi-$1.5$~\cite{xie2024show} & $1.3$B & $84.1$ & $1128.0$ & - & -  & $56.5$ & $30.7$ & - \\
        & MobileVLM~\cite{chu2023mobilevlm} & $1.4$B & $84.5$ & $1196.2$ & $53.2$ & - &  $56.1$ & - & -\\
        & MobileVLM-V2~\cite{chu2024mobilevlm2} & $1.4$B & $84.3$ & $1302.8$ & $57.7$ & - &  $59.3$ & - & -\\
        
        & MobileVLM~\cite{chu2023mobilevlm} & $2.7$B & $84.9$ & $1288.9$ & $59.6$ & - &  $59.0$ & - & -\\
        & MobileVLM-V2~\cite{chu2024mobilevlm2} & $2.7$B & $84.7$ & $1440.5$ & $63.2$ & - &  $61.1$ & - & -\\
        & LLaVA-Phi~\cite{zhu2024llava} & $2.7$B & $85.0$ & $1335.1$ & $59.8$ & -  & - & - & $28.9$\\
        
        & LLaVA~\cite{liu2024visual} & $7$B & $76.3$ & $809.6$ & $38.7$ & $33.5$  & - & - & $25.5$ \\
        & LLaVA-v$1.5$~\cite{liu2024improved}& $7$B & $85.9$ & $1510.7$ & $64.3$ & $58.6$ & $62.0$ & $35.4$ & $31.1$ \\
        & InstructBLIP~\cite{instructblip} & $7$B & - & - & $36.0$ & $53.4$ &  $49.2$ & - & $26.2$ \\
        & Qwen-VL-Chat~\cite{bai2023qwen} & $7$B & - & $1487.5$ & $60.6$ & $58.2$ & $57.5$ & - & - \\
        & IDEFICS-$9$B~\cite{laurencon2023introducing} & $8$B & - & - & $48.2$ & - &  $38.4$ & - & - \\
        & Emu$3$-Chat~\cite{wang2024emu3} & $8$B & $85.2$ & $1244$ & $58.5$ & $68.2$ &  $60.3$ & $31.6$ & $37.2$ \\
        & InstructBLIP~\cite{instructblip} & $13$B & $78.9$ & $1212.8$ & - & - & $49.5$ & - & $25.6$ \\
        
        \midrule
        \textit{Und. and Gen.} 
        & DreamLLM$^\dagger$~\cite{dong2023dreamllm} & $7$B & - & - & - & -  & - & - & $36.6$ \\
        & LaVIT$^\dagger$~\cite{jin2023unified} & $7$B & - & - & - & - & $46.8$ & - & - \\
        & MetaMorph$^\dagger$~\cite{tong2024metamorph} & $8$B & - & - & $75.2$ & $71.8$  & - & - & - \\
        & Emu$^\dagger$~\cite{sun2023generative} & $13$B & - & - & - & - &  - & - & - \\
        & NExT-GPT$^\dagger$~\cite{wu2023next} & $13$B & - & - & - & - &  - & - & - \\
        \cdashline{2-10}
        \\[-2ex]
        & Show-o-256~\cite{xie2024show} & $1.3$B & $73.8$ & $948.4$ & - & - &  $48.7$ & $25.1$ & - \\
        & Show-o-512~\cite{xie2024show} & $1.3$B & $80.0$ & $1097.2$ & - & - &  $58.0$ & $26.7$ & - \\
        & D-Dit~\cite{li2024dual} & $2.0$B & $84.0$ & $1124.7$ & - & - & $59.2$ & - & - \\
        & Gemini-Nano-1~\cite{team2023gemini} & $1.8$B & - & - & - & - &  - & $26.3$ & - \\
        & ILLUME~\cite{wang2024illume} & $7$B &  $88.5$ &  $1445.3$ &  $65.1$ &  $72.9$ &   $-$ & $38.2$ &  $37.0$ \\
        & TokenFlow-XL~\cite{qu2024tokenflow} & $13$B &  $86.8$ &  $1545.9$ &  $68.9$ &  $68.7$ &   $62.7$ & $38.7$ &  $40.7$ \\
        & LWM~\cite{liu2024world} & $7$B & $75.2$ & - & - & - &  $44.8$ & - & $9.6$ \\
        & VILA-U~\cite{wu2024vila} & $7$B & $85.8$ & $1401.8$ & - & $59.0$ &  $60.8$ & - & $33.5$ \\
        & Chameleon~\cite{team2024chameleon} & $7$B & - & - & - & - &  - & $22.4$ & $8.3$ \\
        & Janus & $1.5$B & $87.0$ & $1338.0$ & $69.4$ & $63.7$ &  $59.1$ & $30.5$ & $34.3$ \\
        & \textbf{Janus-Pro-1B} & $1.5$B & $86.2$ & $1444.0$ & $75.5$ & $68.3$ & $59.3$ & $36.3$ & $39.8$ \\
        & \textbf{Janus-Pro-7B} & $7$B & $87.4$ & $1567.1$ & $79.2$ & $72.1$ & $62.0$ & $41.0$ & $50.0$ \\
        
        \bottomrule
    \end{tabular}
\end{table}

\subsection{Evaluation Setup}

\noindent \textbf{Multimodal Understanding.}
To assess multimodal understanding capabilities, we evaluate our model on widely recognized image-based vision-language benchmarks, which include GQA \cite{hudson2019gqa}, POPE \cite{li2023evaluating}, MME \cite{fu2023mme}, SEED \cite{li2023seed}, MMB \cite{liu2023mmbench}, MM-Vet \cite{yu2023mm}, and MMMU \cite{yue2024mmmu}. 

\noindent \textbf{Visual Generation.}
For evaluating visual generation capabilities, we use GenEval~\cite{ghosh2024geneval} and DPG-Bench~\cite{hu2024ella}. GenEval is a challenging benchmark for text-to-image generation, designed to reflect the comprehensive generative abilities of visual generation models by offering a detailed instance-level analysis of their compositional capabilities. DPG-Bench (Dense Prompt Graph Benchmark) is a comprehensive dataset consisting of 1065 lengthy, dense prompts, designed to assess the intricate semantic alignment capabilities of text-to-image models.

\label{sec:evluation_setup}

\subsection{Comparison with State-of-the-arts}

\noindent \textbf{Multimodal Understanding Performance.}
We compare the proposed method with state-of-the-art unified models and understanding-only models in Table~\ref{sota_result_understanding}.
Janus-Pro achieves the overall best results. 
This can be attributed to decoupling the visual encoding for multimodal understanding and generation, mitigating the conflict between these two tasks.
When compared to models with significantly larger sizes,  Janus-Pro remains highly competitive. For instance, Janus-Pro-7B outperforms TokenFlow-XL ($13$B) on all benchmarks except GQA.

\begin{table}[ht]
    \centering
    \setlength{\tabcolsep}{4pt}
    \renewcommand{\arraystretch}{1.2}
    \scriptsize
    \caption{\textbf{Evaluation of text-to-image generation ability on GenEval benchmark}. ``Und.'' and ``Gen.'' denote ``understanding'' and ``generation'', respectively. Models using external pretrained diffusion model are marked with $^\dagger$. 
    }
    \vspace{-2mm}
    \scalebox{0.81}{
    \begin{tabular}{llccccccc}
        \toprule
        \textbf{Type} & \textbf{Method}  & \textbf{Single Obj.} & \textbf{Two Obj.} & \textbf{Counting} & \textbf{Colors} & \textbf{Position} & \textbf{Color Attri.} & \textbf{Overall$\uparrow$} \\
        \midrule
        \multirow{8}{*}{\textit{Gen. Only}} 
        & LlamaGen~\cite{sun2024autoregressive}  & $0.71$ & $0.34$ & $0.21$ & $0.58$ & $0.07$ & $0.04$ & $0.32$ \\
        & LDM~\cite{rombach2022high} & $0.92$ & $0.29$ & $0.23$ & $0.70$ & $0.02$ & $0.05$ & $0.37$ \\
        & SDv$1.5$~\cite{rombach2022high} &  $0.97$ & $0.38$ & $0.35$ & $0.76$ & $0.04$ & $0.06$ & $0.43$ \\
        & PixArt-$\alpha$~\cite{chen2023pixart} &  $0.98$ & $0.50$ & $0.44$ & $0.80$ & $0.08$ & $0.07$ & $0.48$ \\
        & SDv$2.1$~\cite{rombach2022high} &  $0.98$ & $0.51$ & $0.44$ & $0.85$ & $0.07$ & $0.17$ & $0.50$ \\
        & DALL-E $2$~\cite{ramesh2022hierarchical}  & $0.94$ & $0.66$ & $0.49$ & $0.77$ & $0.10$ & $0.19$ & $0.52$ \\
        & Emu$3$-Gen ~\cite{wang2024emu3}  & $0.98$ & $0.71$ & $0.34$ & $0.81$ & $0.17$ & $0.21$ & $0.54$ \\
        & SDXL~\cite{podell2023sdxl} &  $0.98$ & $0.74$ & $0.39$ & $0.85$ & $0.15$ & $0.23$ & $0.55$ \\
        & DALL-E $3$~\cite{dalle3}  & $0.96$ & $0.87$ & $0.47$ & $0.83$ & $0.43$ & $0.45$ & $0.67$ \\
        & SD3-Medium~\cite{esser2024scalingrectifiedflowtransformers} & 0.99 & 0.94 & 0.72 & 0.89 & 0.33 & 0.60 & $0.74$ \\
        \midrule
        \multirow{5}{*}{\textit{Und. and Gen.}}
        & SEED-X$^\dagger$~\cite{ge2024seed}  & $0.97$ & $0.58$ & $0.26$ & $0.80$ & $0.19$ & $0.14$ & $0.49$ \\
        \cdashline{2-9}
        & Show-o~\cite{xie2024show} &  $0.95$ & $0.52$ & $0.49$ & $0.82$ & $0.11$ & $0.28$ & $0.53$ \\
        & D-DiT~\cite{li2024dual} &  $0.97$ & $0.80$ & $0.54$ & $0.76$ & $0.32$ & $0.50$ & $0.65$ \\
        & LWM~\cite{liu2024world} &  $0.93$ & $0.41$ & $0.46$ & $0.79$ & $0.09$ & $0.15$ & $0.47$ \\
        & Transfusion~\cite{zhou2024transfusion} & - & - & - & - & - & - & $0.63$ \\
        & ILLUME~\cite{wang2024illume} &  $0.99$ & $0.86$ & $0.45$ & $0.71$ & $0.39$ & $0.28$ & $0.61$ \\
        & TokenFlow-XL~\cite{liu2024world} &  $0.95$ & $0.60$ & $0.41$ & $0.81$ & $0.16$ & $0.24$ & $0.55$ \\
        & Chameleon~\cite{team2024chameleon} &  - & - & - & - & - & - & $0.39$ \\
        & Janus~\cite{wu2024janus} & $0.97$ & $0.68$ & $0.30$ & $0.84$ & $0.46$ & $0.42$ & $0.61$ \\
        & \textbf{Janus-Pro-1B} &  $0.98$ & $0.82$ & $0.51$ & $0.89$ & $0.65$ & $0.56$ & $0.73$ \\
        & \textbf{Janus-Pro-7B} &  $0.99$ & $0.89$ & $0.59$ & $0.90$ & $0.79$ & $0.66$ & $0.80$ \\
        
        \bottomrule
    \end{tabular}
}
    \label{tab:geneval}
\vspace{-2mm}
\end{table}

\begin{table}[ht]
    \centering
    \renewcommand{\arraystretch}{1.2}
    \scriptsize
    \caption{\textbf{Performances on DPG-Bench.} The methods in this table are all generation-specific models except Janus and Janus-Pro.}
    \vspace{-2mm}
    \scalebox{0.85}{
    \begin{tabular}{lcccccc}
        \toprule
        \textbf{Method} & \textbf{Global} & \textbf{Entity} & \textbf{Attribute} & \textbf{Relation} & \textbf{Other} & \textbf{Overall$\uparrow$} \\
        \midrule
        SDv1.5 \cite{2022LDM} & 74.63 & 74.23 & 75.39 & 73.49 & 67.81 & 63.18 \\
        PixArt-$\alpha$ \cite{chen2023pixart} & 74.97 & 79.32 & 78.60 & 82.57 & 76.96 & 71.11 \\
        Lumina-Next \cite{2024lumina} & 82.82 & 88.65 & 86.44 & 80.53 & 81.82 & 74.63 \\
        SDXL \cite{2023SDXL} & 83.27 & 82.43 & 80.91 & 86.76 & 80.41 & 74.65 \\
        Playground v2.5 \cite{2024PG2.5} & 83.06 & 82.59 & 81.20 & 84.08 & 83.50 & 75.47 \\
        Hunyuan-DiT \cite{2024hunyuandit} & 84.59 & 80.59 & 88.01 & 74.36 & 86.41 & 78.87 \\
        PixArt-$\Sigma$ \cite{2024pixartsigma} & 86.89 & 82.89 & 88.94 & 86.59 & 87.68 & 80.54\\
        Emu3-Gen \cite{wang2024emu3} & 85.21 & 86.68 & 86.84 & 90.22 & 83.15 & 80.60 \\
        DALL-E 3 \cite{dalle3} & 90.97 & 89.61 & 88.39 & 90.58 & 89.83 & 83.50 \\
        SD3-Medium \cite{esser2024scalingrectifiedflowtransformers} & 87.90 & 91.01 & 88.83 & 80.70 & 88.68 & 84.08 \\
        Janus & 82.33 & 87.38 & 87.70 & 85.46 & 86.41 & 79.68 \\
        \textbf{Janus-Pro-1B} & 87.58 & 88.63 & 88.17 & 88.98 & 88.30 & 82.63 \\
        \textbf{Janus-Pro-7B} & 86.90 & 88.90 & 89.40 & 89.32 & 89.48 & 84.19 \\
        \bottomrule
    \end{tabular}
}
    \label{tab:exp-dpg}
    \vspace{-4mm}
\end{table}

\noindent \textbf{Visual Generation Performance.}
We report visual generation performance on GenEval and DPG-Bench. As shown in Table~\ref{tab:geneval}, our Janus-Pro-7B obtains $80$\% overall accuracy on GenEval, which outperforms all the other unified or generation-only methods, e.g., Transfusion~\cite{zhou2024transfusion} ($63$\%) SD3-Medium ($74$\%) and DALL-E $3$ ($67$\%). This demonstrates that our approach has better instruction-following capabilities. As shown in Table~\ref{tab:exp-dpg}, Janus-Pro achieves a score of $84.19$ on DPG-Bench, surpassing all other methods. This demonstrates that Janus-Pro excels in following dense instructions for text-to-image generation.

\vspace{-4mm}
\subsection{Qualitative Results}
\vspace{-3mm}

\begin{figure}[t]
    \centering
    \includegraphics[width=\textwidth]{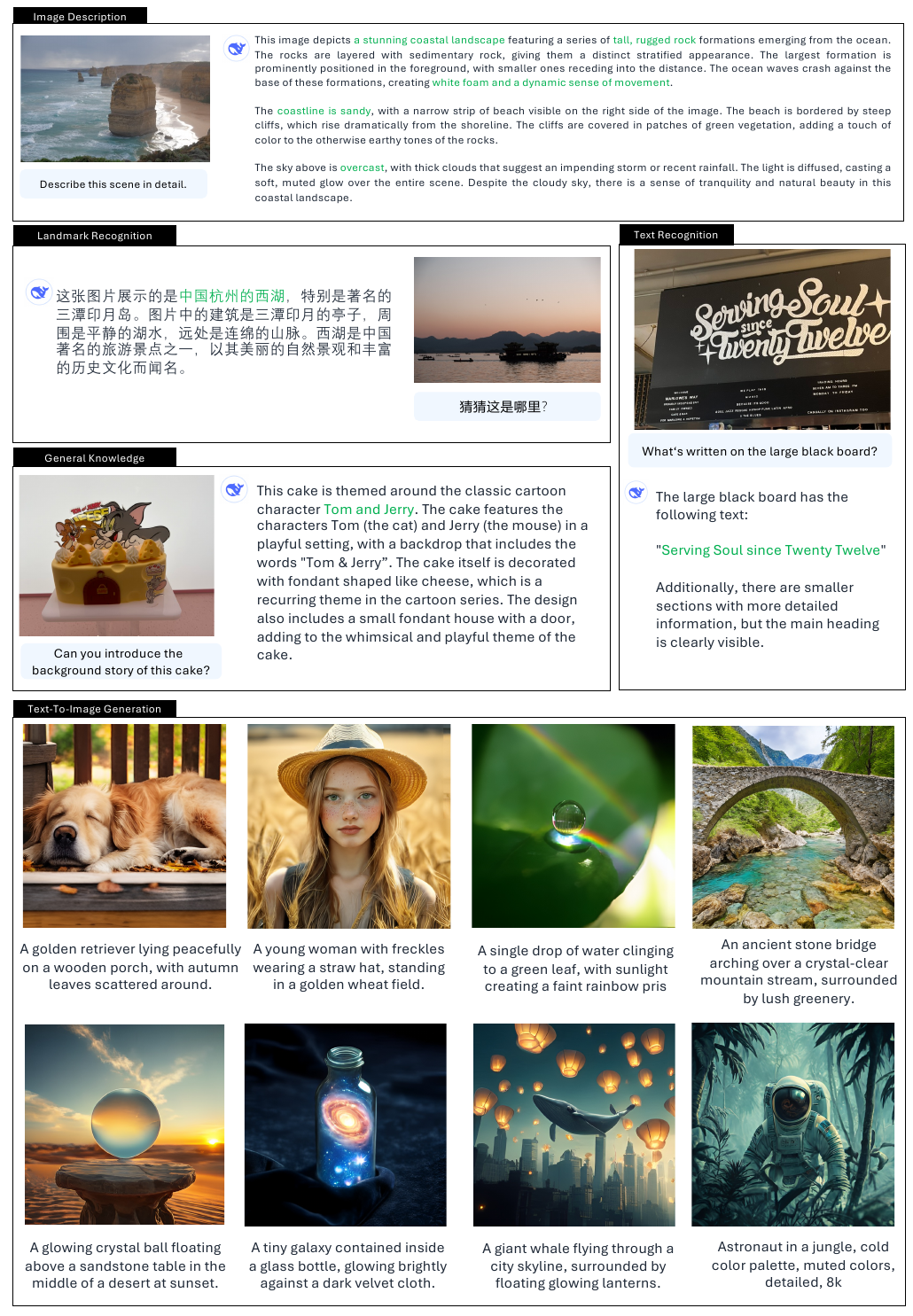}
    \caption{\textbf{Qualitative results of multimodal understanding and visual generation capability}. The model is Janus-Pro-7B and the image output resolution of visual generation is $384 \times 384$. Best viewed on screen.
    } 
    \label{fig:visualization}
\end{figure}

We present results on multimodal understanding in Figure~\ref{fig:visualization}. Janus-Pro exhibits impressive comprehension abilities when handling inputs from various contexts, showcasing its powerful capabilities.
We also present some text-to-image generation results in the lower part of Figure~\ref{fig:visualization}. 
The images generated by Janus-Pro-7B are highly realistic, and despite having a resolution of only $384 \times 384$, they still contain a lot of details. For imaginative and creative scenes, Janus-Pro-7B accurately captures the semantic information from the prompts, producing well-reasoned and coherent images.

\clearpage

\section{Conclusion}
This paper introduces improvements to Janus from three aspects: training strategy, data, and model size. These enhancements have led to significant advancements in both multimodal understanding and text-to-image instruction-following capabilities. However, Janus-Pro still has certain limitations.
In terms of multimodal understanding, the input resolution is limited to 384 $\times$ 384, which affects its performance in fine-grained tasks such as OCR. For text-to-image generation, the low resolution, combined with reconstruction losses introduced by the vision tokenizer, results in images that, while rich in semantic content, still lack fine details. For example, small facial regions occupying limited image space may appear under-detailed. Increasing the image resolution could mitigate these issues.

\bibliographystyle{abbrvnat}
\bibliography{main}

\end{CJK*}
\end{document}